\documentclass[11pt,a4paper]{article}
\usepackage[dvipsnames]{xcolor}
\usepackage[hyperref]{acl2021}

\usepackage{times}
\usepackage{latexsym}

\usepackage{microtype}

\aclfinalcopy 
\usepackage{amsmath}
\usepackage{kantlipsum}
\usepackage{tabularx}
\usepackage[utf8]{inputenc} 
\usepackage[T1]{fontenc}   
\usepackage{hyperref}       
\usepackage{url}
\usepackage{booktabs,multirow} 
\usepackage{amsfonts}       
\usepackage{nicefrac}      
\usepackage{microtype}      
\usepackage{graphicx}
\usepackage{times}
\usepackage{epsfig}
\usepackage{graphicx}
\usepackage{amsmath}
\usepackage{amssymb}
\usepackage{vwcol}

\usepackage{booktabs}

\DeclareMathAlphabet{\pazocal}{OMS}{zplm}{m}{n}

\usepackage{mathtools}
\DeclarePairedDelimiterX{\norm}[1]{\lVert}{\rVert}{#1}
\usepackage{multicol}
\usepackage[linesnumbered,ruled,vlined]{algorithm2e}
\usepackage{setspace}

\let\oldnl\nl
\newcommand{\nonl}{\renewcommand{\nl}{\let\nl\oldnl}}

\usepackage{diagbox}
\DeclareMathAlphabet{\pazocal}{OMS}{zplm}{m}{n} 
\usepackage{microtype}
\usepackage{graphicx}
\usepackage{subfigure}
\usepackage{booktabs} 
\usepackage{times}
\usepackage{latexsym}
\usepackage{makecell} 
\usepackage{amsfonts} 
\usepackage{amsmath} 
\usepackage{amssymb}

\title{Self-Supervised Document Similarity Ranking via Contextualized Language Models and Hierarchical Inference}

\author{Dvir Ginzburg\thanks{\;\;$,\S$\; Denotes equal contribution.}$^{\;\;,1}$~~~Itzik Malkiel$^{*,1,4,}$\thanks{\;\; Corresponding author}~~~Oren Barkan$^{\S,2,4}$~~~Avi Caciularu$^{\S,3}$~~~Noam Koenigstein$^{1,4}$ \\
    $^1$Tel-Aviv University, $^2$Ariel University, $^3$Bar-Ilan University, $^4$Microsoft \\
    {\tt\small \{itmalkie, orenb, Noam.Koenigstein\}@microsoft.com} \\
    {\tt\small \{dvirginz, avi.c33\}@gmail.com} \\
}

\date{}

\begin{document}

\maketitle
\begin{abstract}

We present a novel model for the problem of ranking a collection of documents according to their semantic similarity to a source (query) document. While the problem of document-to-document similarity ranking has been studied, most modern methods are limited to relatively short documents or rely on the existence of ``ground-truth'' similarity labels. Yet, in most common real-world cases, similarity ranking is an unsupervised problem as similarity labels are unavailable. Moreover, an ideal model should not be restricted by documents' length. Hence, we introduce SDR, a self-supervised method for document similarity that can be applied to documents of arbitrary length. 
Importantly, SDR can be effectively applied to extremely long documents, exceeding the $4,096$ maximal token limit of Longformer.
Extensive evaluations on large documents datasets show that SDR significantly outperforms its alternatives across all metrics. To accelerate future research on unlabeled long document similarity ranking, and as an additional contribution to the community, we herein publish two human-annotated test-sets of long documents similarity evaluation. The SDR code and datasets are publicly available \footnote{\href{https://github.com/microsoft/SDR}{github.com/microsoft/SDR}}.

\end{abstract}

\section{Introduction}
Text similarity ranking is an important task in multiple domains, such as information retrieval, recommendations, question answering, and more. Recent approaches based on Transformer language models such as BERT \cite{bert} 
benefit from effective text representations, but are limited in their maximum input text length.
Hence, developing techniques for long-text or document level matching is an emerging research field \cite{47856}. 
In this work, we present SDR, a self-supervised method for document-to-document similarity ranking that can be effectively applied to extremely long documents of arbitrary length and does not require similarity labels.
SDR employs a self-supervised pre-training phase that leverages: (1) a masked language model that fine-tunes \emph{contextual word} embeddings to specialize in a given domain and (2) a contrastive loss on sentence pairs, assembled by inter-and intra-sampling, that encourages the model to produce enhanced text embeddings for similarity. 
Similarity inference is achieved by producing per-sentence embeddings followed by a two-staged hierarchical scoring.

Our contributions are as follows: (1) we present SDR, a novel method for document-to-document similarity that can effectively operate on long documents of arbitrary length and does not require similarity labels. We evaluate SDR and report its performance on two large datasets of documents, showcasing its ability to rank documents better than other state-of-the-art alternatives. (2) to accelerate future research, we publish two long-document similarity datasets annotated by human experts.

\section{Related Work}

\label{sec:related_work}
Semantic similarity has been studied in many fields, such as computer vision~\cite{parmar2018image,huang2017densely}, recommender systems~\cite{ wang-fu-2020-item,barkan2020neural,barkan2021cold,malkiel-etal-2020-recobert}, and natural language processing~\cite{bert,sbert,mikolov2013distributed}. Recently, transformer-based Language Models (LMs) ushered significant performance gains in various natural language understanding tasks, but mainly on relatively short texts \cite{bert, roberta}. These models are usually pre-trained on the Masked Language Modeling (MLM) objective followed by a downstream task-specific fine-tuning process \cite{wang-etal-2018-glue}. 
However, most models employ a battery of self-attention operations, which scale \emph{quadratically} with the sequence length rendering extremely inefficient for long documents containing pages of text. 

To mitigate scale challenges, the Longformer \cite{longformer} model has been proposed which employs local windowed attention unit that restrains the computation and space to scale \emph{linearly} with the sequence length.
However, computation complexity still depends on the sequence length. Moreover, it entails a linear space complexity (memory usage). Therefore, in practice, the propagation of extremely long sequences remains infeasible and the maximal input of the Longformer is capped at $4,096$ tokens only, far less than many real-world long documents. 

Apart from the aforementioned scale limitations on the model's input, in the case of computing pairwise similarities between a large number of documents, the above models also suffer from an exhaustive inference process: 
Longformer and BERT score pairs of items in a unified feed-forward process, by which each pair of two items is fed to the model in order to produce a single pair-wise score (as opposed to scoring based on the individual item embeddings). Such inference technique, impose $O(N^2)$ feed-forward operations~\cite{barkan2020scalable}, compared to just $O(N)$ in SDR.

An additional challenge of Transformer-based LMs is the fact that their raw vector representations is known to perform poorly on semantic textual similarity tasks \cite{sbert}. 
As a result, specific methods for text similarity tasks have been proposed. 
A prominent example for such methods is the SBERT model \cite{sbert}. 
SBERT employs a novel fine-tuning procedure that encourages representations of similar sentences be closer in terms of the cosine similarity, substantially improving their ability to capture semantic similarity. Yet, SBERT still toils from the aforementioned complexity challenges and is unable to handle long-documents.

Recently, several works proposed long-document processing and retrieval techniques using labeled data. \citet{cohan-etal-2020-specter} introduced SPECTER, a model for producing document-level embedding of scientific documents. SPECTER employs a novel objective that uses paper citations as a proxy for similarity. 
Similarly, the Cross-Document Attention (CDA) model \cite{zhou-etal-2020-multilevel} and the Cross-Document Language Model (CDLM) \cite{caciularu2021cross} suggest equipping language models with cross-document information for document-to-document similarity tasks.  
All the above methods rely on supervision, either during the pre-training phase or during fine-tuning. 
However, in the general case, document-to-document similarity (as well as most similarity tasks), is performed in unsupervised settings where no labels (or citations) are available.

\begin{figure}[t]
\centering
\includegraphics[width=1\linewidth]{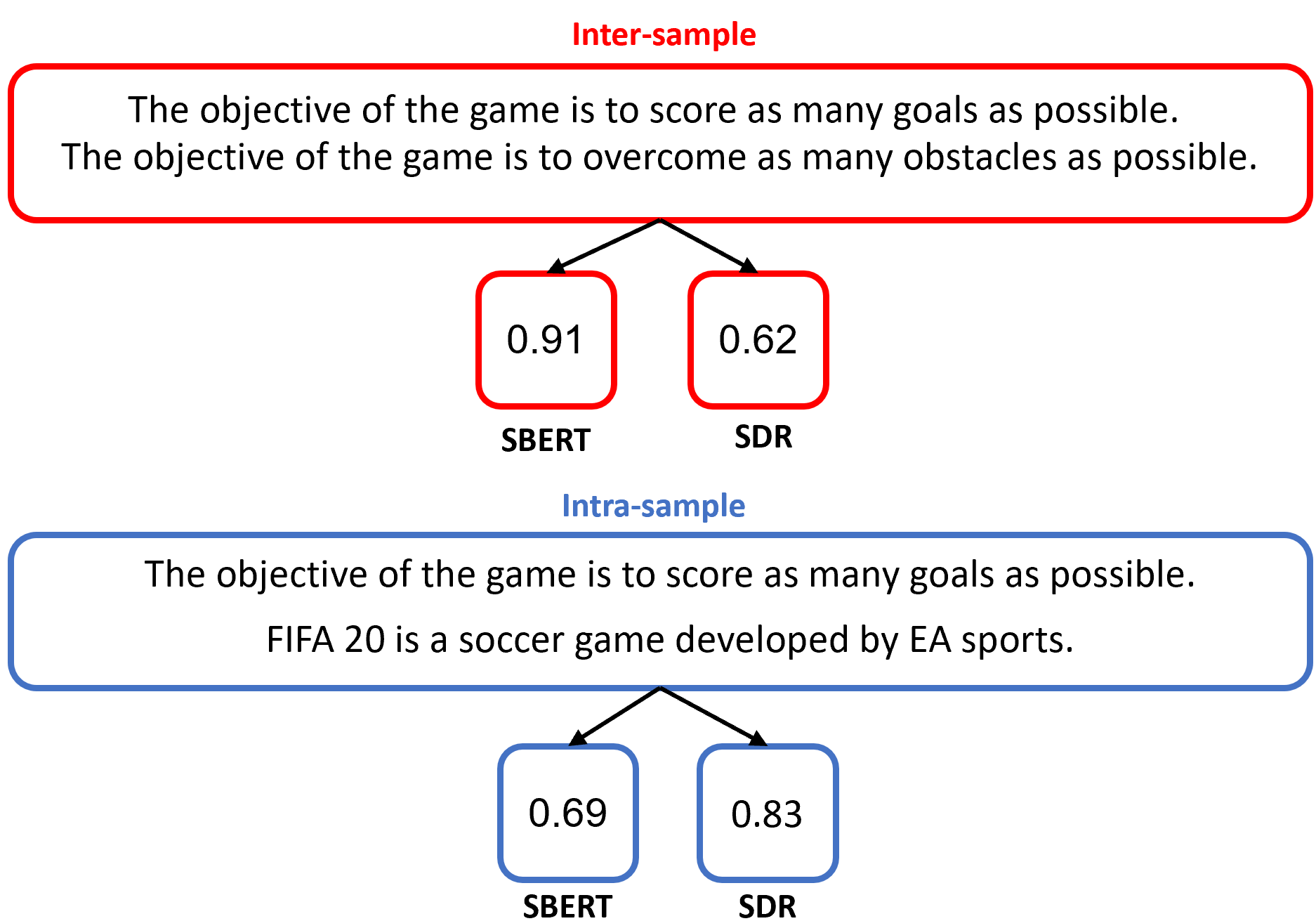}
\caption{
A representative inter- and intra-samples, along with cosine similarity scores retrieved by SBERT and SDR. Top: Inter-sampling from two documents associated with games of different categories. SBERT scores the sentences with a higher cosine value than the one retrieved by SDR. Bottom: attaching the anchor sentence with a sentence sampled from the same paragraph (and document). SDR and SBERT are reversed, where SDR yields a higher score that is more faithful to the sentences' underlying semantics and topic. 
}
\label{Fig:metric_intuition}
\vspace*{-10pt}
\end{figure}

Another line of work consists of hierarchically learning single document representations. 
For example, Hierarchical Attention Networks (HANs) incorporate words and sentences into the final document representation showing competitive performance in different tasks involving long document encoding \cite{yang-etal-2016-hierarchical, sun-etal-2018-stance}. 
More recently, \citet{yang2020beyond} and \citet{47856} investigated hierarchical models based on recurrent neural networks or BERT~\cite{bert} leading to state-of-the-art results in \textit{supervised} document similarity challenges. 
These hierarchical models employ a bottom-up approach in which a long body of text (a document) is represented as an aggregation of smaller components i.e., paragraphs, sentences, and words.
As opposed to these works, SDR exploits a document's hierarchical structure while avoiding compressing it into a single representation. This enables SDR to preserve more relevant information, leading to the superior results presented in Sec.~\ref{sec:experiments}.
Importantly, SDR is \textit{unsupervised} and does not require similarity labels or further fine-tuning.

\begin{figure*}[t]
\centering
\includegraphics[width=1.0\linewidth]{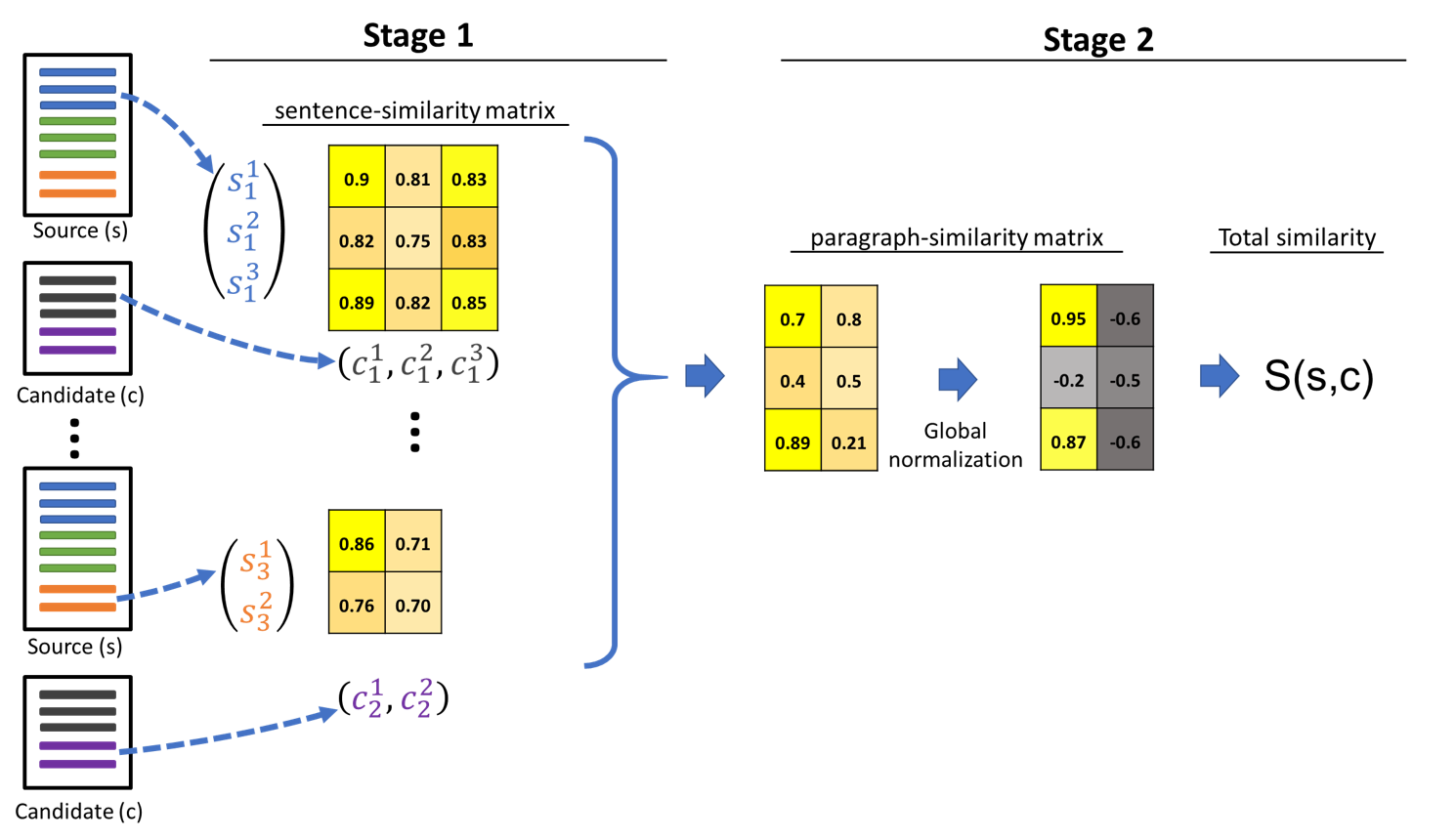}
\caption{A schematic illustration of the SDR inference. Given a source and candidate documents, and for each paragraph-pair, SDR decomposes the paragraphs into sentences and maps each sentence into a vector. In the first stage, a sentence-similarity matrix is computed for each paragraph-pair. In the second stage, paragraph-similarity scores are inferred for all pairs and aggregated into a paragraph-similarity matrix. The matrix is then globally normalized and reduced into a total score, estimating the cumulative similarity between the two documents.}
\label{Fig:inference}
\end{figure*}

\section{The SDR Model}

We present the problem setup followed by a description of the SDR model, its training and  inference.

\subsection{Problem Setup}

Given a collection of documents ${D = \{d_1,  \dots, d_n\}}$ and a source document ${s \in D}$, the goal is to quantify a score that would allow us to rank all the other documents in $D$ according to their semantic similarity with the source document $s$.
In this work, we assume that document similarity labels are not supplied. Therefore, we propose a self-supervision loss that utilizes labels that we invent - this is a proxy to the ultimate similarity labels (if were given).

\subsection{Training} \label{sec:training}
\hypertarget{training}{SDR} adopts the RoBERTa language model as a backbone, and, following \citet{dontstoppretraining}, continues the pre-training of the RoBERTa model on $D$. Unlike RoBERTa, the SDR training solely relies on negative and positive sentence-pairs produced by inter- and intra-document sampling, respectively.

Specifically, the SDR training propagates sentence-pairs sampled from $D$. The sentence-pairs are sampled from the same paragraph 
with probability 0.5 (intra-samples), otherwise from different paragraphs taken from the different documents (inter-samples). The sentences in each pair are then tokenized, 
aggregated into batches, and randomly masked in a similar way to the RoBERTa pre-training paradigm. 
The SDR objective comprises a dual-term loss. The first term is the standard MLM loss adopted from ~\citet{bert}. 
Denoted by $\mathcal{L}_{MLM}$. The MLM loss allows the model to specialize in the domain of the given collection of documents~\cite{gururangan-etal-2020-dont}.

\hypertarget{contrastive}{The} second loss term is the contrastive loss~\cite{hadsell2006dimensionality}. Given a sentence pair $(p,q)$ propagated through the model, we compute a feature vector for each sentence by average pooling the token embeddings associated with each sentence separately. The tokens embedding are the output of the last encoder layer of the model. The contrastive loss is then applied to the pair of feature vectors and aims to encourage the representations of intra-samples to become closer to each other while pushing inter-samples further away than a predefined positive margin $m \in R^{+}$. 

Formally, the contrastive loss is defined as follows:
\begin{equation}
\mathcal{L}_{C}=\left\{\begin{aligned}1-C\left(f_{p}, f_{q}\right) \;\;\; & y_{p, q}=1 \\
\max \left(0, C\left(f_{p}, f_{q}\right)-(1-m)\right) \;\;\; & y_{p, q}=0
\end{aligned}\right.
\end{equation}
where $f_p, f_q$ are the pooled vectors extracted from the tokens embedding of sentence $p$ and $q$, respectively. $y(p,q) = 1$ indicates an intra-sample (sentence-pair sampled from the same paragraph), otherwise negative (sentence-pair from different documents). $C(f_p, f_q)$ measures the angular distance between $f_p$ and $f_q$ using the Cosine function:
\begin{equation}
C\left(f_p, f_q\right)=\frac{f_p^{T} f_q}{\left|f_p\right|\left|f_q\right|}
\end{equation}
A demonstration of the inter-and intra-sampling procedure associated with the cosine scores produced by SDR can be found in Fig.\ref{Fig:metric_intuition}. 
The figure presents a representative sample as a motivation for SDR sampling and contrastive loss, where SDR is shown to score sentences in a way that is more faithful to their underlying topic and semantics.
Importantly, as the inter-samples represent sentences that were randomly sampled from different documents, it is not guaranteed that their semantics would oppose each other. Instead, it is likely that those sentences are semantically uncorrelated while obtaining some level of opposite semantics only in rare cases. Therefore, instead of pushing negative samples to completely opposite directions, we leverage the contrastive loss in a way that encourages orthogonality between inter-samples while avoiding penalizing samples with negative scores. Hence, in our experiments, we set $m \triangleq 1$, which encourages inter-samples to have a cosine similarity that is less than or equal to $0$, and do not penalize pairs with negative cosine scores.

Finally, both loss terms are combined together yielding the total loss
\begin{equation}
\mathcal{L}_{total}= \mathcal{L}_{MLM} + 
\mathcal{L}_{C}
\end{equation}

\pagebreak
\subsection{Inference}\label{sec:inference}
\hypertarget{inference}{Let} $s \in D$ be a source document composed of a sequence of paragraphs $s=(s_i)_{i=1}^{\widetilde{n}}$, where each paragraph comprises a sequence of sentences $s_i=(s_i^k)_{k=1}^{i^*}$, and $i^*$ denotes the number of sentences in $s_i$. Similarly, let $c \in D$ be a candidate document, $c$ can be written as $c=(c_j)_{j=1}^m$, where $c_j=(c_j^r)_{r=1}^{j^*}$. 
The SDR inference scores the similarity between $s$ and every other candidate document $c$ by calculating two-staged hierarchical similarity scores. The first stage operates on sentences to score the similarity between paragraph-pairs, and the second operates on paragraphs to infer the similarity between two documents. 
In SDR, we first map each document in $D$ into a sequence of vectors by propagating its sentences through the model. Each sentence is then transformed into a vector by average pooling the token embeddings of the last encoder layers’ outputs.
Next, for each candidate document $c \in D$, SDR iterates over the feature vectors associated with the sentences in $s$ and $c$ and composes a \emph{sentence-similarity matrix} for each paragraph-pair from both documents. Specifically, for each paragraph-pair $(s_i, c_j) \in s \times c$, SDR computes the cosine similarity between every pair of sentence embedding from $s_i \times c_j$, forming a sentence-similarity matrix. Focusing on the $(k,r)$ cell of this matrix, $1 \leq k \leq i^*$, $1 \leq r \leq j^*$, the sentence-similarity matrix can be expressed as:
\begin{equation}
M_{ij}^{kr} \triangleq C(s_i^k, c_j^r)
\end{equation}
Calculated for each paragraph pair $(s_i, c_j) \in s \times c$, the paragraph-similarity scores are then aggregated into a \emph{paragraph-similarity matrix}. Focusing on the $(i,j)$ cell, the matrix can be expressed as:

\begin{equation}
\label{eq:paragraph-score}
P_{ij}^{sc}\triangleq \frac{\sum_{k=1}^{i^*} \max\limits_{0 \leq r \leq j^*} M_{ij}^{kr}}{i^*}
\end{equation}
The motivation behind the similarity scores in Eq.~\ref{eq:paragraph-score} is that similar paragraph-pairs should incorporate similar sentences that are more likely to correlate under the cosine metric, due to the properties of the contrastive loss employed throughout SDR training. 
In order to rank all the documents in the dataset, we compute the above paragraph-similarity matrix for every candidate document $c \in D$. \hypertarget{normalization}{The} resulted paragraph-similarity matrices are then globally normalized. 
Each row $i$ in $P_{ij}^{sc}$ is z-score normalized by a mean and standard deviation computed from the row $i$ values of $P_{ij}^{sc}$ across all candidates $c \in D$. The motivation behind this global normalization is to refine the similarity scores by highlighting the ones of the most similar paragraph-pairs and negatively scores the rest. 
Throughout our early experiments, we observed that different paragraph-pairs incorporate sentences with different distributions of cosine scores, where some source paragraphs may yield a distribution of cosine values with a sizeable margin compared to other paragraphs. This can be attributed to the embedding space, for which some regions can be denser than others.

Finally, a total similarity score is inferred for each candidate $c$, using the above paragraph-similarity matrix. The total similarity score aims to quantify the cumulative similarity between $s$ and $c$. To this end, we aggregate all paragraph-similarity scores for each paragraph in $s$ as follows:

\begin{equation}
\label{eq:total_score}
\mathcal{S}(s,c) = \frac{\sum_{i=1}^{\widetilde{n}} \max\limits_{1 \leq j \leq m}  \left[ \textrm{NRM} (P_{ij}^{sc})\right]_{i,j}} {n}
\end{equation}

where NRM is the global normalization explained above. 
The essence of Eq.\ref{eq:total_score} is to match between the most similar paragraphs from $s$ and $c$, letting those most correlated paragraph-pairs contribute to the total similarity score between both documents. Finally, the ranking of the entire collection $d$ can be obtained by sorting all candidate documents according to $\mathcal{S}(s,c)$, in a descending order. 

It is important to notice that (1) in SDR inference, we do not propagate documents-pairs through the language model (which is computationally exhaustive). Instead, the documents are separately propagated through the model. Then, the scoring solely requires applications of non-parametric operations\footnote{The cosine similarity function.}. (2) both SDR training and inference operate on sentences and therefore do not suffer from discrepancies between the two phases.

\section{Experiments}
\label{sec:experiments}

\begin{figure*}[t]
\centering
\includegraphics[width=1.0\linewidth]{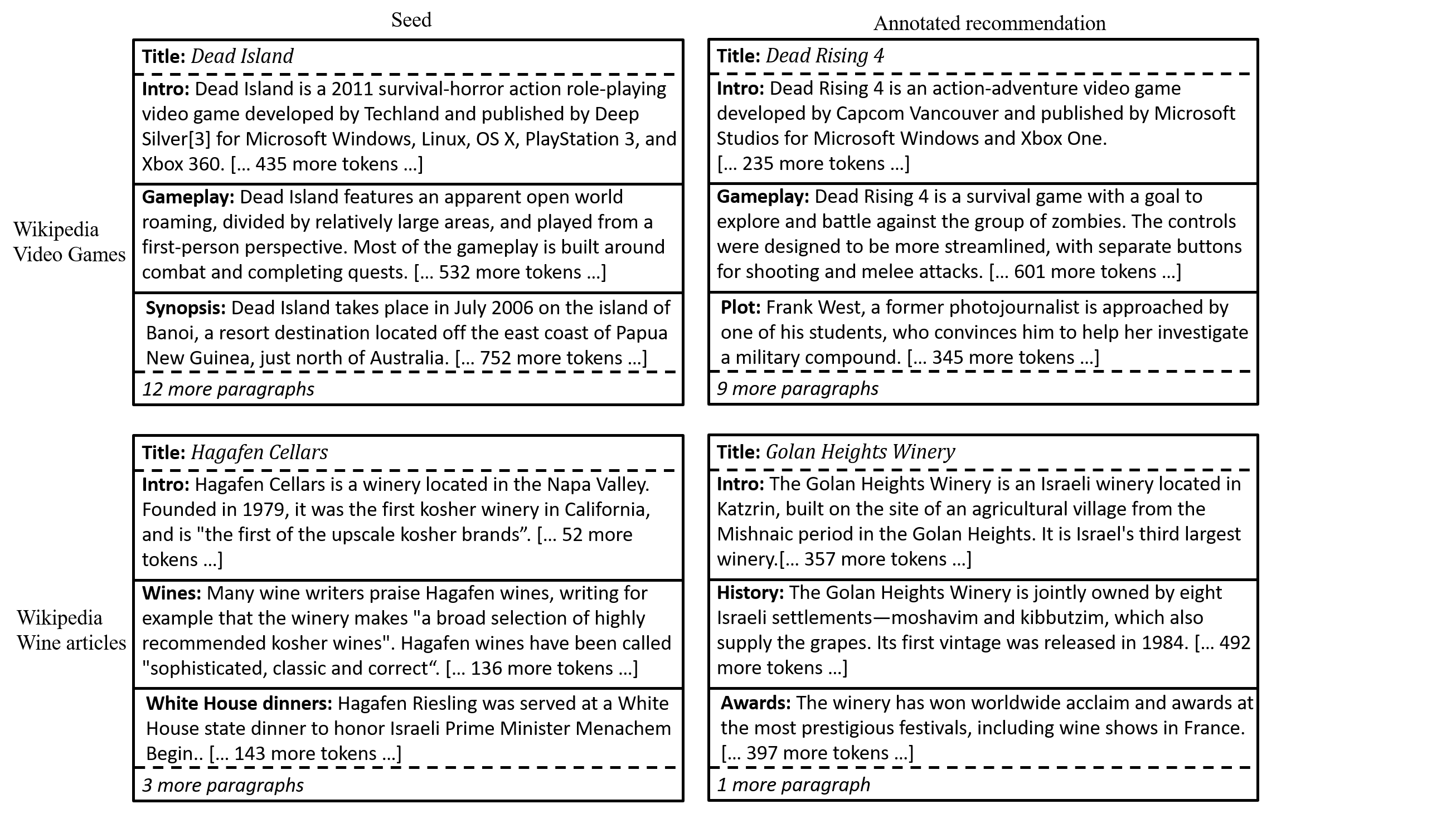}
\caption{Samples from the Wikipedia Video Games (WVG) and Wikipedia Wines Articles (WWA) datasets. For each seed item (left), the opening sentence of each of the first three paragraphs is presented. A recommended sample by the domain expert is shown on the right.}
\label{Fig:datasets_samples}
\end{figure*}

\subsection{Datasets} \label{sub:datasets}

We conducted our experiments over two datasets excerpted from Wikipedia. For each of the Wikipedia-based datasets, we provide a human-annotated test set of similarity labels. Examples from the datasets are provided in Fig. \ref{Fig:datasets_samples}.  

\paragraph{Wikipedia video games (WVG)}

The Wikipedia video games dataset\footnote{\href{https://zenodo.org/record/4812962\#.YK5z2KgzY2w}{Wikipedia Video Games dataset.}} consists of 21,935 articles reviewing video games from all genres and consoles. Each article consists of a different combination of sections, such as summary, gameplay, plot, production, etc. For this dataset, we publish ground-truth similarity annotations, crafted by a domain expert, for $\sim 90$ source game articles. For each source, the expert annotated $\sim 12$ articles of similar games. Examples for the ground-truth similarities are: (1) Grand Theft Auto - Mafia, (2) Burnout Paradise - Forza Horizon 3.

\paragraph{Wikipedia wine articles (WWA)} Wikipedia wines\footnote{\href{https://zenodo.org/record/4812960\#.YK5z2agzY2w}{Wikipedia Wine Articles dataset.}} dataset consists of 1635 articles from the wine domain. This dataset consists of a mixture of articles discussing different types of wine categories, brands, wineries, grape varieties, and more. The ground-truth similarities were crafted by a human sommelier who annotated 92 source articles with ${\scriptsize \sim}10$ similar articles, per source. Examples for ground-truth expert-based similarities are: (1) Dom Pérignon - Moët \& Chandon, (2) Pinot Meunier - Chardonnay.

\subsection{Quantitative Metrics}\label{sec:quantitative_metrics}

We evaluated the performance of SDR, the baselines, and ablations using the MPR,MRR and HR@$k$ metrics:
\paragraph{Mean Percentile Rank (MPR)}
The mean percentile rank is the average of the percentile ranks for every sample with ground truth similarities in the dataset. Given a sample $s$, the percentile rank for a true recommendation $r$ is the rank the model gave to $r$ divided by the number of samples in the dataset. 
MPR evaluates the stability of the model, i.e, only models where all ground truth similarities had a high rank by the model will have a good score. 
\vspace{-6pt}
\paragraph{Mean Reciprocal Rank (MRR)} The mean reciprocal rank is the average of the best reciprocal ranks for every sample with ground truth similarities in the dataset. Given a sample with $M_s$ ground truth similarities we first mark the rank of each ground truth recommendation by the model and then take the reciprocal of the \emph{best} (lowest) rank.
\vspace{-6pt}
\paragraph{Hit Ratio at $k$ (HR@$k$)} 
HR@$k$ evaluates the percentage of true predictions in the top $k$ retrievals made by the model, where a true prediction corresponds a candidate sample from the ground truth annotations.

\begin{table*}[t!]
\centering
\begin{tabular}{@{}l@{~}l@{~}c@{~}c@{~}c@{~}c@{~}c@{~}c@{~}c@{~}c@{~}c@{~}c@{~}c@{~}c@{}}
\toprule
&& \multicolumn{4}{c}{Video games}& \multicolumn{4}{c}{Wines}\\
\cmidrule(lr){3-6}
\cmidrule(lr){7-10}
\textbf{Architecture}& \textbf{Inference}&  \textbf{
\  MPR\ \ }&  \textbf{\ MRR\ \ }&
\textbf{HR@10}& \textbf{HR@100}&  \textbf{\  MPR\ \ }&  \textbf{\ MRR\ \ } &\textbf{HR@10}& \textbf{HR@100}
\\
\midrule

LDA& - 
& 94.1\%   & 31.8\%     & 8.8\%  & 28.1\%
& 83.7\%   & 23.4\%     & 8.7\%  & 41.3\%\\
 
\noalign{\vskip 2mm}  
  \bottomrule 
  \noalign{\vskip 2mm} 
  
 SBERT& First 
& 86.4\%&42.6\%&11.9\%&26.9\%
&83.5\%&31.1\%&11.4\%&41.8\% \\
  SBERT& ALL 
&92.6\%&51.1\%&16.1\%&37.5\% 
&81.3\%&28.3\%&11.1\%&37.2\% \\
  SBERT& SDR\textsubscript{inf}  
& 94.2\%& 53.4\%&18.2\%&39.7\% 
&83.6\%&32.3\%&12.1\%&41.0\% \\

\noalign{\vskip 2mm}  
  \bottomrule 
  \noalign{\vskip 2mm}

 Longformer& CLS 
&58.0\%&10.4\%&2.1\%&6.3\%  
&65.0\%&15.7\%&4.8\%&14.6\% \\
 Longformer& First 
&66.6\%&3.3\%&1.3\%&3.7\%  
&56.1\%&12.4\%&3.5\%&10.2\% 
 \\
  Longformer& ALL 
&66.0\%&9.7\%&2.4\%&4.3\%   
&64.7\%&13.9\%&2.2\%&11.8\%\\
  Longformer& SDR\textsubscript{inf} 
&68.5\%&10.2\%&4.1\%&7.7\%  
&55.3\%&16.4\%&3.3\%&12.3\% \\
 
\noalign{\vskip 2mm}  
  \bottomrule 
  \noalign{\vskip 2mm}    
 BERT large& CLS 
&69.6\%&30.9\%&9.7\%&20.3\%  
&70.1\%&30.3\%&9.5\%&34.7\% \\
 BERT large& First 
&61.1\%&15.5\%&3.8\%&9.8\%  
&71.3\%&21.8\%&6.7\%&20.9\% \\
  BERT large& ALL 
&65.2\%&27.2\%&7.1\%&16.2\%  
&64.6\%&27.8\%&7.8\%&26.2\% \\
  BERT large& SDR\textsubscript{inf} 
&71.2\%&33.5\%&12.9\%&22.2\%  
&75.5\%&24.7\%&9.7\%&36.8\% \\

\noalign{\vskip 2mm}  
  \bottomrule 
  \noalign{\vskip 2mm} 
  
  SDR& SDR\textsubscript{inf} 
&\textbf{97.4\%}&\textbf{64.0\%}&\textbf{23.6\%}&\textbf{54.0\%}  
&\textbf{89.3\%}&\textbf{50.9\%}&\textbf{17.0\%}&\textbf{59.0\%} \\
 
\bottomrule
\end{tabular}
\caption{Similarity results evaluated on the video games (left), movies (middle) and wines (right) datasets from wikipedia, based on expert annotations. 
The second column specifies the applied inference method, as described in Section \ref{inference_methods}. SBERT\textsubscript{\large v} refers to the vanilla SBERT (without continuing training on each dataset by utilizing our pseudo-labels). }
\label{Tab:results}
\end{table*}









\subsection{Baseline models}
\label{sec:baselines}

We compare SDR with the following baselines:
\paragraph{Latent Dirichlet Allocation (LDA)} LDA \cite{lda} is one of the renowned algorithms for topic modeling and text-matching. 
LDA assumes that documents are generated by sampling from a distribution of latent topics, where each topic can be described by another distribution defined over the vocabulary.
For every LDA experiment, we perform a grid search with $1,000$ different configuration of hyper-parameters. The reported performance corresponds to the model with the highest topic coherence value \cite{topiccoherence}.

\paragraph{BERT and Longformer} 
For BERT \cite{bert} and Longformer \cite{longformer} (see Sec.~\ref{sec:related_work}), we evaluate two different variants. First, we employ the publicly available pre-trained weights of the models. Second, we continue the pre-training of the models over the corpora induced by the datasets, applying our proposed method associated with each model.  
We used only the ``large'' architectures during all the experiments.

\paragraph{SBERT}

The SBERT model~\cite{sbert} utilizes a fine-tuning approach that produces semantically meaningful embeddings under a cosine-similarity metric. 

We evaluate SBERT model under two inference configurations (1) using the original weights trained on the NLI dataset \cite{nli}, and (2) after fine-tuning with the pseudo labels presented in Sec. \ref{sec:training}.

\subsection{Inference Methods}\label{sec:inf_base_models}
To compare SDR with the above baselines, which are restricted by a maximal sequence length, 
we follow previous procedures~\cite{sbert, longformer} and report the performance of four different inference techniques applied on the output embeddings of the different models :

\begin{itemize}
  \item \textbf{CLS} - use the special CLS token embedding of the $N$\footnote{$N$ is the maximal sequence length the model supports in one forward pass.} first tokens.
  \item \textbf{FIRST} - use the mean of the embeddings of the $N$ first tokens.
  \item \textbf{ALL} - propagating the entire document in chunks, then use the mean of the embeddings of all the tokens in the sample.
  \item \textbf{SDR\textsubscript{inf}} - use the hierarchical SDR inference described in Sec.~\ref{sec:inference}.
\end{itemize}

\label{inference_methods}

\begin{table*}[t]
\centering
\resizebox{1.\textwidth}{!}{%
\begin{tabular}{ccccc}
\toprule
& \multicolumn{2}{c}{Video games}& \multicolumn{2}{c}{Wines}\\
\cmidrule(lr){2-3}
\cmidrule(lr){4-5}

\backslashbox{Model}{Seed}  & Dead Island  & Mafia III    & Hagafen Cellars & Champagne                                     

 \\ \hline
\multirow{3}{*}{SDR}   & \multicolumn{1}{l}{1.   {\color{ForestGreen}Dead Island: Riptide}}  & \multicolumn{1}{l}{1. {\color{ForestGreen}Mafia II}}     &  \multicolumn{1}{l}{1.   {\color{ForestGreen}Golan Heights Winery}} & \multicolumn{1}{l}{1. {\color{ForestGreen}Champagne Krug}} \\
                       & \multicolumn{1}{l}{2. {\color{ForestGreen}Dying Light}}                    & \multicolumn{1}{l}{2. {\color{ForestGreen}Saints Row 2}}      & \multicolumn{1}{l}{2. {\color{ForestGreen}Manischewitz}}                    & \multicolumn{1}{l}{2. {\color{ForestGreen}Sparkling wine} }        \\
                       & \multicolumn{1}{l}{3. {\color{ForestGreen}Dead Rising 4}}                                              & \multicolumn{1}{l}{3. {\color{ForestGreen}Grand Theft Auto V}} & \multicolumn{1}{l}{3. {\color{ForestGreen}Barkan Wine Cellars}}                                              & \multicolumn{1}{l}{3. {\color{ForestGreen}Champagne Krug}} \\ \hline

\multirow{3}{*}{SBERT} & \multicolumn{1}{l}{1. {\color{Goldenrod}Fallout 3}} & \multicolumn{1}{l}{1. {\color{Goldenrod}Red Dead Redemption 2}}   & \multicolumn{1}{l}{1.   {\color{BrickRed}Petit Rouge}} & \multicolumn{1}{l}{1. {\color{ForestGreen}Moët \& Chandon}}         \\
                       & \multicolumn{1}{l}{2. {\color{ForestGreen}Dead Rising 3}}                    & \multicolumn{1}{l}{2. {\color{Goldenrod}Dark Souls II}}     & \multicolumn{1}{l}{2. {\color{Goldenrod}Roter Veltliner}}                    & \multicolumn{1}{l}{2. {\color{Goldenrod}Chardonnay}}                      \\
                       & \multicolumn{1}{l}{3. {\color{Goldenrod}Wasteland 2}}                                              & \multicolumn{1}{l}{3. {\color{BrickRed}Battlefield} }  & \multicolumn{1}{l}{3. {\color{ForestGreen}Trisaetum Winery}}                        & \multicolumn{1}{l}{3. {\color{BrickRed}Chasselas}} \\ \hline

\multirow{3}{*}{BERT} &  \multicolumn{1}{l}{1.   {\color{Goldenrod}The Outer Worlds}} & \multicolumn{1}{l}{1. {\color{ForestGreen}The Godfather}}   &  \multicolumn{1}{l}{1.   {\color{Goldenrod}Domaine Dujac}} & \multicolumn{1}{l}{1. {\color{Goldenrod}Roter Veltliner}}                     \\
                       & \multicolumn{1}{l}{2.  {\color{ForestGreen}Metro Exodus}}                    & \multicolumn{1}{l}{2. {\color{BrickRed}Dark Souls}}       & \multicolumn{1}{l}{2. {\color{ForestGreen}Petri Wine}}                    & \multicolumn{1}{l}{2. {\color{BrickRed}Table wine}}                   \\
                       & \multicolumn{1}{l}{3. {\color{Goldenrod}Rage 2}}                                              & \multicolumn{1}{l}{3. {\color{Goldenrod}Code Vein}} & \multicolumn{1}{l}{3. {\color{BrickRed}Blue Nun}}                                              & \multicolumn{1}{l}{3. {\color{ForestGreen}Champagne wine region}} \\ \bottomrule

\end{tabular}%
}
\caption{
Similarity predictions for the Wikipedia video games (WVG) and Wikipedia wine articles (WWA) datasets. For each of the shown recommendations, a domain expert rated the similarity with the source document. Red, yellow, and green indicate poor, mediocre, and high similarity (respectively).}

\label{tab:qualititive_results}
\end{table*}

\subsection{Results}

The results over the document similarity benchmarks are depicted in Tab.~\ref{Tab:results}. The scores are based on the ground-truth expert annotations associated with each dataset. The results indicate that SDR outperforms all other models by a sizeable margin. 
Recall that the underlying LMs we evaluated (BERT, Longformer) were pre-trained on the MLM objective. This makes them hard to generate meaningful embeddings suitable for probing similarity using the Cosine-similarity metric, as previously discussed in Sec.~\ref{sec:related_work}. Comparing to the best variant of each model, SDR presents absolute improvements of $\sim$7-12\% and $\sim$11-13\% in MPR, and MRR, respectively, and across all datasets. 

SBERT, as opposed to the underlying models above, presents a cosine similarity-based loss during training. Compared to SDR, we observe that a fine-tuned SBERT, which utilizes the pseudo-labels introduced in Sec. \ref{sec:training}, shows inferior results across all datasets, yielding -3\% MPR, -5\% MRR and -2\% HR@$10$ in the Video games. This can be attributed to SBERT's cosine loss, that constantly penalizes negative pairs to reach a cosine score of $-1$. For uncorrelated sentence-pairs, such property can hinder the convergence of the model. See the below ablation analysis for more details. 
We observe that SBERT's suffers from an additional degradation in performance when applied with the original SBERT weights, yielding -6\% MPR and -8\% MRR. This can be attributed to the importance of continue training on the given dataset at hand.

Notably, as shown in the table, applying the SDR inference to other baseline language models improves their performance by a significant margin. This is another evidence of our inference's advantage over other methods, especially as we observe sizeable gains across all baseline models and datasets. 

Inspecting SBERT results, we see that the SDR\textsubscript{inf} gains increase in all metrics, yielding an increase of at least +3\% MPR, +4\% MRR, +6\% HR@$10$ and +7\% HR@$100$. This can be attributed to the importance of the hierarchical evaluation for long documents and indicate the struggle transformers have in embedding long text into a single vector.   
Importantly, SDR outperforms SBERT by a significant margin, even when SBERT is applied with SDR\textsubscript{inf}. This is due to SDR training, which incorporates the contrastive loss for promoting orthogonality between negative sentence-pairs.

Table~\ref{tab:qualititive_results} presents qualitative results on randomly chosen samples from the WWA and WVG datasets. We compare SDR with the top two baselines associated with the highest scores in the Wikipedia evaluations, namely SBERT and BERT.
Similarly to the evaluation scheme presented for the quantitative experiments, we employ a self-supervised training for SBERT, with the same pseudo-labels as in SDR training. 
In Tab.~\ref{tab:qualititive_results}, we observe that SDR correctly understands the essence of the article, as finding \textit{Grand Theft Auto V} similar to \textit{Mafia III}, or \textit{Sparkling wine} similar to \textit{Champagne}.
As to SBERT results, we see that the model fails to grasp the article's underlying topic in 50\% of the predictions. For example, SBERT matches between \textit{Battlefield} and \textit{Mafia III}, or \textit{Chasselas} and \textit{Champagne}. This can be attributed to the fact that SBERT does not apply a hierarchical inference and struggles to compress the entire document representation in one vector. This becomes especially crucial in very long documents, which are common in the WVG dataset.
In BERT, we observe document similarity predictions of relatively poor quality. For example, for \textit{Hagafen Cellars} BERT retrieves \textit{Blue Nun}, or for \textit{Champagne} it matches \textit{Table wine}. 
The relative degradation in performance can be attributed to the BERT pre-training procedure, which inherently does not optimize text embedding under a well-defined metric. 

The above results highlight the benefit obtained by the SDR model, which utilizes a hierarchical inference, along with a self-supervised training procedure that embeds sentences under a well-defined similarity metric.

\subsection{Ablation Study}
\label{ablation}

\begin{table*}[t!]
\centering
\begin{tabular}{@{}l@{~}l@{~}c@{~}c@{~}c@{~}c@{~}c@{~}c@{~}c@{~}}
\toprule
&&\multicolumn{3}{c}{Video games}\\
\cmidrule(lr){3-5}
&&\textbf{MPR}&\textbf{MRR}&
\textbf{HR@10}  \\
\midrule

(i)&No hierarchical inference&96.3\%
&52.4\%
&20.1\%\\

 (ii)&Paragraph-level inference&97.4\%
&58.5\%
&22.8\%\\

(iii)&No training&87.1\%
&28.2\%
&7.0\%\\

(iv)&No normalization&97.3\%
&63.2\%
&22.6\%\\

(v)&No contrastive loss&91.5\%&46.0\%&14.5\%\\

&Full method&\textbf{97.4\%}&\textbf{64.0\%}&\textbf{23.6\%}\\
\bottomrule
\end{tabular}
\caption{Ablation study results. 
}
\label{Tab:ablation}
\end{table*}

We performed an ablation study to asses the effectiveness of SDR. To that end, we used the video games dataset, described in Sec. \ref{sub:datasets}. The following ablations are considered: 

\begin{itemize}
\item \hyperlink{inference}{No hierarchical inference} - the embeddings of the first $N$ tokens of each document are averaged, producing one embedding vector per document. These embeddings are compared via the cosine function to score the similarity between documents. This is similar to the scoring procedure from \cite{sbert}.
\item \hyperlink{inference}{Paragraph-level inference} - the paragraph-similarity matrix is computed directly using the first $N$ tokens of each paragraph. This variant neglects the sentence-similarity matrix from stage 1 \ref{Fig:inference} of the inference mechanism. The scoring proceeds by stage 2 of the inference, as described in Sec.~\ref{sec:inference}.
\item \hyperlink{training}{No training} - the BERT pre-trained weights are used and applied with the proposed hierarchical inference (i.e., we do not employ additional pre-training on the given collection of documents).
\item \hyperlink{normalization}{Global normalization} - the SDR inference is applied without globally normalizing the paragraph-similarity matrix. 
\item \hyperlink{contrastive}{No contrastive loss} - the SDR training is applied without the contrastive loss term (solely using the MLM objective).
\item \hyperlink{contrastive}{Standard cosine loss} - the SDR training employs a contrastive loss with a margin of $m=2$. This is equivalent to the standard Cosine-Similarity loss, that reinforces negative and positive samples to cosine scores of $-1$ and $1$, respectively.
\end{itemize}

The results depicted in Tab.\ref{Tab:ablation} indicate that our proposed hierarchical inference is highly beneficial, even compared to a paragraph-level inference, that it is crucial to employ the proposed training in the way it is done in SDR, and that it is better to apply global normalization.

Particularly noticeable is the contrastive loss, whose gain is present in both (ii) and (iii), for which the biggest degradation in the results took place. Another significant improvement is due to the hierarchical inference, with a leap of 11\% in MPR by applying paragraph-level inference, and another 9\% by applying the two-stage hierarchy.

\subsection{Implementation details}
SDR and all other transformer-based baselines utilize the Huggingface package \footnote{\href{https://huggingface.co/}{HuggingFace}}. In our transformer-based baselines experiments, we use the best-published model configuration associated with each variant. To split the paragraphs into sentences as suggested in SDR, we used the NLTK package \footnote{\href{https://www.nltk.org/}{nltk}}, resulting in an average sentence length of 16 tokens.
We use a train-validation split of 90\%-10\% to evaluate the MLM and cosine similarity accuracy during training.

For LDA modeling and similarity evaluation, we used the implementation of the Gensim package\footnote{\href{https://radimrehurek.com/gensim/}{Gensim}}. We conduct a hyperparameter search, based on the topic coherence score, to find the best LDA parameters for each dataset.

For SBERT we used the official package\footnote{\href{https://www.sbert.net/}{SBERT API}}, with the released fine-tuned weights for the STS task.
All our experiments were conducted using a single Tesla V100 32GB card, with a batch size of 8 both for training and evaluation.

\section{Conclusions}

In this work, we presented Self-Supervised Document Similarity Ranking (SDR), a novel self-supervised model for document similarity, supporting extremely long documents.
Documents' similarities are extracted via a hierarchical bottom-up scoring procedure, which preserves more semantic information, leading to superior similarity results. 
For our evaluations, we assembled two manually-labeled test-sets using expert annotations, that will be made publicly available to expedite future 
research on long-document similarities.

\bibliography{acl2021}
\bibliographystyle{acl_natbib}

\end{document}